\def\year{2018}\relax

\documentclass[letterpaper]{article} 
\usepackage{aaai18}  
\usepackage{times}  
\usepackage{helvet}  
\usepackage{courier}  
\usepackage{url}  
\usepackage{graphicx}  
\usepackage{dsfont}
\usepackage[inline]{enumitem}
\usepackage{researchpack}
\usepackage{xcolor}

\newcommand{\cp}{\textsc{cp}}
\newcommand{\vbeus}{\textsc{vb-eus}}
\newcommand{\vbqi}{\textsc{vb-qi}}
\newcommand{\setmargin}{\textsc{SetMargin}}

\begin{document}
%
\title{Constructive Preference Elicitation over Hybrid Combinatorial Spaces}
\author{Paolo Dragone\thanks{PD is a fellow of TIM-SKIL Trento and is supported by a TIM scholarship.}\\
University of Trento, Italy\\
TIM-SKIL, Trento, Italy\\
\texttt{paolo.dragone@unitn.it}
\And
Stefano Teso\thanks{This work has received funding from the European Research
Council (ERC) under the European Union’s Horizon 2020 research and innovation
programme (grant agreement No [694980] SYNTH: Synthesising Inductive Data
Models).}\\
KU Leuven, Belgium\\
\texttt{stefano.teso@cs.kuleuven.be}
\And
Andrea Passerini\\
University of Trento, Italy\\
\texttt{andrea.passerini@unitn.it}
}

\maketitle

\begin{abstract}
    Preference elicitation is the task of suggesting a highly preferred
    configuration to a decision maker. The preferences are typically learned by
    querying the user for choice feedback over pairs or sets of objects.  In
    its \emph{constructive variant}, new objects are synthesized ``from
    scratch'' by maximizing an estimate of the user utility over a
    combinatorial (possibly infinite) space of candidates.
    In the constructive setting, most existing elicitation techniques fail
    because they rely on exhaustive enumeration of the candidates. A previous
    solution explicitly designed for constructive tasks comes with no formal
    performance guarantees, and can be very expensive in (or unapplicable to)
    problems with non-Boolean attributes.
    We propose the \textit{Choice Perceptron}, a Perceptron-like algorithm for
    learning user preferences from set-wise choice feedback over constructive
    domains and hybrid Boolean-numeric feature spaces. We provide a theoretical
    analysis on the attained regret that holds for a large class of
    query selection strategies, and devise a heuristic strategy that aims at
    optimizing the regret in practice.
    Finally, we demonstrate its effectiveness by empirical evaluation against
    existing competitors on constructive scenarios of increasing complexity.
\end{abstract}

\section{Introduction}

Constructive preference elicitation is the task of recommending structured
objects, i.e. configurations of several components, assembled on the basis of the
user preferences~\cite{teso2016constructive,dragone2017layout}. In this setting,
the space of possible configurations grows exponentially in the number of
components. Examples include configurable products, such as personal computers
or mobile phone plans, and complex preference-based decision problems, such as
customized travel planning or personalized activity scheduling.

The suggested configurations should reflect the preferences of the user, which
are unobserved and must be estimated. As in standard preference
elicitation~\cite{pigozzi16preferences}, preferences can be learned by
iteratively suggesting candidate products to the user, and refining an estimate
of the preference model from the received feedback.  The ultimate goal is to
produce good recommendations with minimal user effort. Here we focus on
\emph{choice queries}, an interaction protocol consisting in recommending a
\emph{set} of products; the user is invited to indicate the most preferred item
in the set~\cite{viappiani2011recommendation,louviere2000stated}.  Elicitation
techniques based on choice set queries rely on some strategy to select
the next query set to show to the user. Successful query selection strategies
must balance between the estimated informativeness of the recommendations (so to
minimize the number of elicitation rounds) and their quality (to maximize the
chance of the user buying the product and to keep her engaged). By generalizing
pairwise ranking feedback, choice queries over larger sets of items allow finer
control over informativeness, diversity and
quality~\cite{pu2009user,bollen2010understanding}.

Most existing preference elicitation methods are not designed for constructive
tasks~\cite{viappiani2011recommendation,teso2016constructive}. Regret-based
methods~\cite{viappiani2009regret} rely on perfectly rational user responses,
while Bayesian approaches do not scale to combinatorial product
spaces~\cite{viappiani2010optimal}, as discussed in the related work section.
A notable exception is the approach of Teso et al.~\cite{teso2016constructive},
which avoids the enumeration of the product space by encoding it through
mixed-integer linear constraints. Alas, it requires configurations to be
encoded with binary variables (in one-hot format), which can be very costly
from a computational perspective, and comes with no formal performance
guarantees.

In this paper we present several contributions. First, we propose an iterative
algorithm, dubbed \emph{Choice Perceptron}, that generalizes the structured
Perceptron~\cite{collins2002discriminative,shivaswamy2015coactive} to
interactive preference elicitation from pairwise and set-wise choice feedback.
The query selection strategy is implemented as an optimization problem over the
combinatorial space of products. In contrast to previous constructive
approaches~\cite{teso2016constructive}, our algorithm handles general linear
utilities over \emph{arbitrary} feature spaces, including combinatorial and
numerical attributes and features. Second, we prove that under
a very general assumption (implied by many existing user response models), the
expected average regret suffered by our algorithm decreases at least as
$\calO(1 / \sqrt{T})$. We show how the constants appearing in the bound
depend on intuitive properties of the query selection strategy, and, as a third
contribution, we propose a simple strategy to control these quantities.
Our empirical analysis showcases the
effectiveness of our approach against several state-of-the-art (including
constructive) alternatives.

\section{Related work}

Preference elicitation (PE) is a widely studied subject in
AI~\cite{domshlak2011preferences,pigozzi16preferences}.
Most existing approaches to PE rely on regret
theory~\cite{viappiani2009regret,viappiani2013robust} or Bayesian
estimation~\cite{viappiani2010optimal}; see~\cite{viappiani2011recommendation}
for a brief overview. None of them are suitable for constructive
settings, for different reasons. Regret-based methods maintain a version space
of utility functions consistent with the collected feedback. However,
inconsistent user responses, which are common in real-world recommendation,
make the version space collapse. Bayesian methods gracefully deal with
inconsistent feedback by employing a full distribution over the candidate
utility functions. Unfortunately, selection of the query set (based on
optimizing its Expected Value of Information or approximations thereof) is
computationally expensive, preventing these approaches from scaling to larger
combinatorial domains.

The only approach specifically designed for constructive preference elicitation is
\setmargin, introduced in~\cite{teso2016constructive}. \setmargin\ can be seen
as a max-margin approximation of Bayesian methods that maintains only $k$ most
promising candidate utility functions (with $k$ small, e.g. $2$ to $4$). Like
the Choice Perceptron, it avoids the explicit enumeration of the product
catalogue by compactly defining the latter in terms of MILP constraints, for
significant runtime benefits.  Alas, it only handles configurations encoded in
one-hot form, which can become inefficient for very complex problems involving
many categorical variables, relies on a rather involved optimization problem,
and it has not be analyzed from a theoretical standpoint. Our query strategy is
much simpler, and aims specifically at optimizing an upper bound on the regret.

Our method is related to Coactive
Learning~\cite{shivaswamy2015coactive}, which has already found application in
constructive tasks~\cite{teso2017coactive,dragone2017layout}; some concepts and
arguments used in our theoretical analysis are adapted from the Coactive
Learning literature~\cite{shivaswamy2012online,raman2013stable}.  However, in
our framework the user is asked to choose an option from a set of
alternatives, rather than to construct an improved configuration. The two
approaches are complementary in the sense that when manipulative feedback is
easy to obtain Coactive Learning may be better suited; however when the space
of products is highly constrained, producing feasible improvements may be
difficult for the user, and our approach is preferable.


\begin{algorithm}[t]
\caption{\label{alg:perp} The Choice Perceptron (\cp) algorithm.}
\begin{algorithmic}[1]
    \Procedure{\cp\;}{$T, \eta$}
        \State $\vw^1 \gets 0$
        \For{$t = 1, \ldots, T$}
            \State Receive context $x^t$ from the user
            \State $\calQ^t \gets \textsc{SelectQuery}(x^t, \vw^t)$
            \State User chooses $\bar{y}^t$ from $\calQ^t$
            \State $\vw^{t+1} \gets \vw^t + \eta \Delta^t$
        \EndFor
    \EndProcedure
\end{algorithmic}
\end{algorithm}

\section{The Choice Perceptron algorithm}

We consider a combinatorial space $\calY$ of structured products defined by
hard feasibility constraints.
As customary in preference elicitation, we focus on the problem of learning a
\textit{utility function} that ranks candidate objects according to the user
preferences.
The utility of a product $y\!\in\!\calY$
may optionally depend on some externally provided context $x\!\in\!\calX$.
In the rest of the paper, we assume that the user's true utility function is
fixed and never observed by the algorithm, and that it
is linear, i.e. of the form $u^*(x, y) = \langle \vw^*, \vphi(x,
y) \rangle$; here $\vw^*\!\in\!\bbR^d$ are the true preference weights of the
user and $\vphi : \calX\!\times\!\calY \rightarrow \bbR^d$ maps context-configuration
pairs to a $d$-dimensional feature space. The
feature vectors $\vphi(x, y)$ are assumed to be enclosed in a
ball of radius $R$.

We propose the Choice Perceptron (\cp) algorithm; the pseudocode is listed in
Algorithm~\ref{alg:perp}.
The \cp\ algorithm keeps an estimate $u^t(x, y) = \langle \vw^t, \vphi(x, y)
\rangle$ of the true user utility, and iteratively refines it by interacting
with the user.
At each iteration $t$, the algorithm receives a context $x^t$ and recommends a
set of $k$ configurations $\calQ^t\!\subseteq\!\calY$, by selecting them
according to some query strategy based on $\vw^t$\footnote{The
\cp\ algorithm is independent from the particular query selection strategy
used. Different query strategies may find better recommendations in different
problems.}. After receiving the query set, the user chooses the ``best'' object
$\bar{y}^t\!\in\!\calQ^t$ according to her preferences.
This kind of set-wise interaction protocol generalizes pairwise ranking
feedback, and is well studied in decision theory, psychology, and
econometrics~\cite{louviere2000stated,toubia2004polyhedral,pu2009user}.
We allow the choice to be noisy, i.e. the user may choose
$\bar{y}^t$ according to a
distribution
$P_{x^t}(\bar{y}^t \! = \! y | y \in \calQ^t)$.

After observing the user's pick, the algorithm updates the current estimate
$\vw^t$. Here we focus on the following Perceptron update:
\begin{align}
    & \vw^{t+1} \gets \vw^t + \eta \Delta^t \nonumber \\
    & \Delta^t := \vphi(x^t, \bar{y}^t) - \frac{1}{k-1}\sum_{y\in\calQ^t : y\neq\bar{y}^t} \vphi(x^t, y) \label{eq:update}
\end{align}
where $\eta$ is a constant step-size. Despite its simplicity, this update comes
with sound theoretical guarantees, as shown in the next section\footnote{Further,
our results could be extended to more sophisticated updating mechanisms, see
e.g.~\cite{shivaswamy2015coactive}.}.

We measure the quality of a recommendation set $\calQ^t$ in context $x^t$ by
the \textit{instantaneous regret}, that is the difference in true
utility between a truly optimal object $y^*_{x^t} \!=\!  \argmax_{y\in\calY}
u^*(x^t, y)$ and the best option in the set:
$$ \reg(x^t, \calQ^t) = \min_{y\in\calQ^t} \left( u^*(x^t, y^*_{x^t}) - u^*(x^t, y) \right) $$
This definition is in line with previous works on preference elicitation with
set-wise choice feedback~\cite{viappiani2010optimal}.  After $T$ iterations, the
\textit{average regret} is $\reg^T\!=\!  \frac{1}{T}\sum_{t=1}^T
\reg(x^t, \calQ^t)$.
A low average regret implies low instantaneous regret throughout the
elicitation process, as is necessary for keeping the user engaged.
In the next section we prove a theoretical upper bound on the expected average
regret suffered by \cp\ under a very general assumption on the user feedback.


\section{Theoretical Analysis}
In this section we analyze the theoretical properties of the \cp\ algorithm,
proving an $\calO(1/\sqrt{T})$ upper bound on its expected average regret.
In the following $\bbE_{\bar{y}^t}[f(x^t, y) | \calQ^t]$ indicates the
conditional expectation of $f(x^t, y)$ with respect to $P_{x^t}(\bar{y}^t \! =
\! y | \calQ^t)$, where $t$ is the iteration index; $\bbE[f(x^t, y)]$ is the
expectation of $f(x^t, y)$ over the distribution of all user choices
$\bar{y}^1, \ldots, \bar{y}^t$.  We will also use the shorthands $P_t(y) :=
P_{x^t}(\bar{y}^t\!=\!y|y\in\calQ^t)$, $u^*(\Delta^t) := \langle \vw^*, \Delta^t
\rangle$ and $[k] := \{1, \dots, k\}$.

In order to derive the regret bound, we need to quantify the ``quality'' of the
sets provided by the query strategy. To this end, we adapt the concept of
expected $\alpha$-informativeness from the Coactive Learning
framework~\cite{shivaswamy2012online}:
\begin{defn}
\label{def:alpha}
For any query strategy, there exist $\alpha\in(0, 1]$ and $\bar{\xi}^t\in\bbR$
such that, for all $t\in[T]$ and for all users:
\begin{align}
    & \bbE_{\bar{y}^t}[u^*(\Delta^t)|\calQ^t] \ge \nonumber \\
    & \qquad \qquad \alpha \max_{y\in\calQ^t} \left( u^*(x^t, y^*_{x^t}) - u^*(x^t, y) \right) - \bar{\xi}^t \label{eq:alphainf-xi}
\end{align}
\end{defn}
The LHS of Eq.~\ref{eq:alphainf-xi} is
the expected \textit{utility gain} of the update rule (Eq.~\ref{eq:update}): a
positive utility gain indicates that $\vw^{t+1}$ makes a step towards a better
approximation of $\vw^*$. The term $\max_{y\in\calQ^t} \left( u^*(x^t,
y^*_{x^t}) - u^*(x^t, y) \right)$ on the RHS is instead the \textit{worst-case
regret}, i.e. the regret with respect to the worst object in the query set. This
model simply quantifies the amount of utility gain in terms of a fraction
$\alpha$ of the worst-case regret and the slack term $\bar{\xi}^t$. Intuitively,
$\alpha$ captures the minimum quality of the query sets selected by the query
strategy, while the slacks $\bar{\xi}^t$ are additional degrees of freedom that
depend on the expected user replies.

Notice that the above definition is very general and can describe the behavior
of any query selection strategy, provided appropriate values for $\alpha$ and
$\bar{\xi}^t$. Both occur as constants in our regret bound.

By requiring the user to behave ``reasonably'', according to the following
definition, we can guarantee the expected utility gain to always be non-negative
(Lemma~\ref{th:ugaingeqzero}). This allows us to make explicit and assign a
precise meaning to the value of the constant $\bar{\xi}^t$.
\begin{defn}
\label{def:reasonable}
A user is \emph{reasonable} if, for any context $x^t$ and
query set $\calQ^t$, the probability $P_{x^t}(\bar{y}^t \! = \! y | \calQ^t)$
is a non-decreasing monotonic transformation of the true utility $u^*$:
$$ \forall y, y' \! \in \! \calQ^t \quad P_t(y) \ge P_t(y') \iff u^*(x^t, y) \ge u^*(x^t, y') $$
\end{defn}
\noindent
This property is implied by many widespread user response models, including the
Bradley-Terry~\cite{bradleyterry} and
Thurstone-Mosteller~\cite{mcfadden2001economic} models of pairwise choice
feedback, and the Plackett-Luce~\cite{plackett1975analysis,luce1959individual}
model of set-wise choice feedback. It is also strictly less restrictive than
applying any of these models.

Notably, when applied to a reasonable user, the update rule
(Eq.~\ref{eq:update}) \emph{always} yields a non-negative expected utility gain.
\begin{lem} \label{th:ugaingeqzero}
For a reasonable user with utility $u^*$, it holds that
$\bbE_{\bar{y}^t}[u^*(\Delta^t)|\calQ^t] \geq 0 $ at all iterations $t$.
\begin{proof}
Given that the user is reasonable, we apply the Chebyshev's sum inequality to
$u^*(x^t, y^t)$ and $P_{x^t}(\bar{y}^t\!=\! y^t | \calQ^t)$, for $y^t\in\calQ^t$:
\begin{align*}
    & \textstyle \frac{1}{k} \sum_{y^t\in\calQ^t} u^*(x^t, y^t) P_t(y^t) \ge \\
    & \textstyle \qquad \left(\frac{1}{k}\sum_{y^t\in\calQ^t} u^*(x^t, y^t) \right)\cdot\left(\frac{1}{k}\sum_{y^t\in\calQ^t} P_t(y^t) \right) \\
    \Longleftrightarrow \quad & \textstyle \sum_{y^t\in\calQ^t} u^*(x^t, y^t) P_t(y^t) \ge \frac{1}{k}\sum_{y^t\in\calQ^t} u^*(x^t, y^t)
\end{align*}
Rearranging, we obtain:
\begin{align*}
    & \textstyle \sum_{y^t\in\calQ^t} u^*(x^t, y^t) P_t(y^t) - \frac{1}{k}\sum_{y^t\in\calQ^t} u^*(x^t, y^t) \ge 0 \\
    \Longleftrightarrow & \textstyle \quad \frac{k}{k-1} \bbE[ u^*(x^t, \bar{y}^t) | \calQ^t] - \frac{1}{k-1}\sum_{y^t\in\calQ^t} u^*(x^t, y^t) \ge 0 \\
    \Longleftrightarrow & \textstyle \quad \bbE[ \frac{k}{k-1} u^*(x^t, \bar{y}^t) - \frac{1}{k-1}\sum_{y^t\in\calQ^t} u^*(x^t, y^t) | \calQ^t] \ge 0 \\
    \Longleftrightarrow & \textstyle \quad \bbE[ \frac{k - 1}{k-1} u^*(x^t, \bar{y}^t) - \frac{1}{k-1}\sum_{y^t\neq\bar{y}^t} u^*(x^t, y^t) | \calQ^t] \ge 0 \\
    \Longleftrightarrow & \textstyle \quad \bbE[ u^*(x^t, \bar{y}^t) - \frac{1}{k-1}\sum_{y^t\neq\bar{y}^t} u^*(x^t, y^t) | \calQ^t] \ge 0
\end{align*}

\end{proof}
\end{lem}
The lemma allows us to distinguish between \textit{informative} and
\textit{uninformative} query sets, depending on whether the expected utility
gain is strictly positive or null, respectively.
We can use these definitions to derive an equivalent formulation of the
$\alpha$-informativeness making the constants $\bar{\xi}^t$ explicit.

Let $\alpha > 0$ be the smallest constant such that
$\bbE_{\bar{y}^t}[u^*(\Delta^t) | \calQ^t]\!\ge\!  \alpha \max_{y\in\calQ^t}
\left( u^*(x^t, y^*_{x^t}) - u^*(x^t, y) \right)$ for all iterations $t$ in
which the query set $\calQ^t$ is informative. For these iterations setting
$\bar{\xi}^t = 0$ still satisfies the inequality in Eq.~\ref{eq:alphainf-xi}.
On the other hand, when the query set is uninformative, $\bar{\xi}^t$ must
satisfy $\bar{\xi}^t \!\ge\! \alpha \max_{y\in\calQ^t} \left( u^*(x^t,
y^*_{x^t}) - u^*(x^t, y) \right)$. Given that $\norm{\vphi(x, y)} \le R$, the
worst-case regret is upper-bounded by $2R\norm{\vw^*}$, therefore it suffice to
set $\bar{\xi}^t \!=\! 2 \alpha R \norm{\vw^*}$. We can rewrite the expected
$\alpha$-informativeness as:
\begin{align}
    &\bbE_{\bar{y}^t}[u^*(\Delta^t) | \calQ^t] \ge \nonumber \\
    &\quad \alpha \max_{y\in\calQ^t} \left( u^*(x^t, y^*_{x^t}) - u^*(x^t, y) \right) - 2 \alpha R \norm{\vw^*} m^t \label{eq:alphainf}
\end{align}
\noindent
Here $m^t\!=\!\mathds{1}[\bbE[u^*(\Delta^t)]\!=\!0]$ is a constant that is
equal to $1$ if \emph{any} query set $\calQ^t$ that may be chosen at iteration
$t$ is expected to be uninformative and $0$ otherwise. Note that
$\bbE[u^*(\Delta^t)] = \bbE[\bbE_{\bar{y}^t}[u^*(\Delta^t) | \calQ^t]]$,
therefore if $\calQ^t$ is informative then $\bbE[u^*(\Delta^t)] > 0$
(i.e. $m^t\!=\!0$), while if $\calQ^t$ is uninformative then $\bbE[u^*(\Delta^t)] = 0$
(i.e. $m^t\!=\!1$). We say that an iteration $t$ is \textit{expected uninformative}
if $m^t \!=\! 1$, and let $M\!:=\!\sum_{t=1}^T m^t$ be the total number of
expected uninformative iterations.

The last property of the query selection strategy we define in order to state the
bound is the $\beta$-affirmativeness, which we adapt from~\cite{raman2013stable}
as follows:
\begin{defn}
For any query selection strategy and for a fixed time horizon $T$, there exists a
constant $\beta\in\bbR$ such that $\frac{1}{T}\sum_{t=1}^T \bbE[u^t(\Delta^t)]
\leq \beta$.
\end{defn}
\noindent
This definition states that $\beta$ is an upper bound on the average expected
change in $u^t$, for $t\in[T]$. Notice that $\bbE[u^t(\Delta^t)]$ may be
positive, null or negative. Intuitively, a negative $\bbE[u^t(\Delta^t)]$
indicates that the query set is expected to produce a user choice that disagrees
with the current estimate of $\vw^t$. This is the case in which the algorithm
receives the most information. In general, the smaller $\beta$ is, the quicker
\cp\ learns from the user feedback.

The previous assumptions on the user and definitions for the query strategy
allow us to derive the following regret bound for \cp\ along the same lines of
what done in Coactive Learning~\cite{shivaswamy2012online,raman2013stable}.
\begin{thm}
\label{th:regb}
For a reasonable user with true preference weights $\vw^*$ and an
$\alpha$-informative and $\beta$-affirmative query strategy, the expected
average regret of the \cp\ algorithm is upper bounded by:
$$ \bbE[\reg^T] \leq \frac{\sqrt{2\frac{\beta}{\eta} + 4R^2} \| \vw^* \|}{\alpha \sqrt{T}} + \frac{2 R \norm{\vw^*} M }{T} $$
\begin{proof}
Using Cauchy-Schwarz and Jensen's inequalities:
\begin{align}
    \bbE[\langle \vw^*, \vw^{T+1} \rangle]
        &\leq \| \vw^* \| \; \bbE[\| \vw^{T+1} \|] \nonumber \\
        &\leq \| \vw^* \| \; \sqrt{\bbE[\langle \vw^{T+1}, \vw^{T+1} \rangle]} \label{eq:jensen}
\end{align}
From the expected $\beta$-affirmativeness and $\norm{\vphi(x, y)} \leq R$:
\begin{align}
    &\bbE[\langle \vw^{T+1}, \vw^{T+1} \rangle] = \nonumber \\
    &\quad = \bbE[\langle \vw^T, \vw^T \rangle] + 2 \eta \bbE[\langle \vw^T, \Delta^T \rangle] + \eta^2 \bbE[ \langle \Delta^T, \Delta^T \rangle ] \nonumber \\
    &\quad \leq 2 \eta \sum_{t=1}^T \bbE[\langle \vw^t, \Delta^t \rangle] + 4 \eta^2 R^2T \leq 2 \eta \beta T + 4 \eta^2 R^2T \nonumber
\end{align}
Plugging this result into inequality (\ref{eq:jensen}) we have:
$$ \bbE[\langle \vw^*, \vw^{T+1} \rangle] \leq \sqrt{2 \eta \beta T + 4 \eta^2 R^2T} \| \vw^* \| $$
For a reasonable user, the $\alpha$-informativeness in Eq.~\ref{eq:alphainf}
holds for any query strategy. Applying it to the LHS of the above inequality,
along with the law of total expectation, we get:
\begin{align*}
          & \textstyle \bbE[\langle \vw^*, \vw^{T+1} \rangle] \\
      = \ & \textstyle \bbE[\langle \vw^*, \vw^T \rangle] + \eta \bbE[\bbE_{\bar{y}^T}[\langle \vw^*, \Delta^T \rangle | \calQ^t]] \\
      = \ & \textstyle \bbE[\sum_{t=1}^{T} \eta \bbE_{\bar{y}^t}[u^*(\Delta^t) | \calQ^t]] \\
\end{align*}
Applying the $\alpha$-informativeness (Eq.~\ref{eq:alphainf}):
\begin{align*}
    \ge \ & \textstyle \alpha \eta \bbE[\sum_{t=1}^{T} \max_{y\in\calQ^t} \left( u^*(x^t, y^*_{x^t}) - u^*(x^t, y) \right)] \\
          & \textstyle \qquad - 2R\alpha\eta\norm{\vw^*}\sum_{t=1}^{T} m^t \\
    \ge \ & \textstyle \alpha \eta \bbE[\sum_{t=1}^{T} \min_{y\in\calQ^t} \left( u^*(x^t, y^*_{x^t}) - u^*(x^t, y) \right)] \\
          & \textstyle \qquad - 2R\alpha\eta\norm{\vw^*}\sum_{t=1}^{T} m^t \\
      = \ & \textstyle \alpha \eta T \bbE[\reg^T] - 2R\alpha\eta\norm{\vw^*}M
\end{align*}
Finally:
\begin{align*}
          & \alpha \eta T \bbE[\reg^T] \\
    \le \ & \sqrt{2 \eta \beta + 4 \eta^2 R^2} \| \vw^* \| \sqrt{T} + 2R\alpha\eta\norm{\vw^*} M
\end{align*}
from which the claim follows.
\end{proof}
\end{thm}

\section{Query selection strategy}

In the previous section we proved an upper bound on the expected
average regret of \cp\ for any query selection strategy, provided that
the user is reasonable.  Crucially, however, the bound depends on the
actual value of $\alpha$, $\beta$ and $M$. These constants depend both
on the user and the query selection strategy. While the algorithm has
no control on the user, an appropriate design of the query selection
strategy can positively affect the impact of the constants on the
bound.  In the following we present a query selection strategy that
aims at reducing the bound by finding a trade-off between $\alpha$ and
$\beta$.

Recall that we want $\alpha \in (0, 1]$ to be large and $\beta \in \bbR$ and $M
\in [T]$ small. While have no direct control over $\alpha$ and $\beta$, which
depend on all iterations, we can control their step-wise surrogates:
\begin{align*}
    \textstyle u^*(\Delta^t) &= \textstyle \langle \vw^*, \Delta^t \rangle \propto \|\vw^*\| \|\Delta^t\| \;\; \text{for $\alpha$} \\
    \textstyle u^t(\Delta^t) &= \textstyle \langle \vw^t, \Delta^t \rangle \propto \|\vw^t\| \|\Delta^t\| \;\; \text{for $\beta$}
\end{align*}
There is a \textit{trade-off} between the two, as they both
depend on $\|\Delta^t\|$. Further, while $\vw^t$ is observed, $\vw^*$ is not.
We proceed as follows. Since $\vw^*$ is not observed, we indirectly maximize
$u^*(\Delta^t)$ by maximizing $\|\Delta^t\|$, i.e. by picking $k$ query
configurations that are distant in feature space.
For reasonable users, maximizing the distance between objects also tends to
maximize the probability $P_t(y)$ of picking a high utility object: the
larger the distance, the higher the probability of picking objects with large
difference in $u^*(\cdot)$. On the other hand $\vw^t$ is observed, so we
can choose $k$ query configurations with small difference in estimated
utility $u^t(\cdot)$ by taking them from a plane orthogonal (or almost
orthogonal) to $\vw^t$. This way, $u^t(\Delta^t)$ is close to $0$ regardless of
the choice of the user, implying $\beta \approx 0$. This reasoning leads to the
following optimization problem:

\begin{align*}
    \calQ^t = \argmax_{\{y_1, \dots, y_k\}} & \quad \gamma \delta + (1 - \gamma) \mu \\
    \text{s.t.} \quad  & \textstyle u^t(x^t, y_1)\! =\! \max_y u^t(x^t, y) \\
                        & \textstyle \vphi(x^t,y_1) \neq \dots \neq \vphi(x^t,y_k) \\
    \text{where:} \quad & \textstyle \delta := \sum_{i = 2}^k \norm{\vphi(x^t, y_1) - \vphi(x^t, y_i)}_1 \\
                        & \textstyle \mu    := \sum_{i = 2}^k u^t(x^t, y_j)
\end{align*}

The objective aims at optimizing a convex combination of the $L_1$ distances of
the options in $\calQ$ ($\delta$) and their distance from optimality ($\mu$).
The two terms are modulated by the $\gamma \in [0, 1]$ hyperparameter.
The third constraint forces the first configuration $y_1$ to
be optimal, irrespective of the choice of $\gamma$, ensuring that when $\vw^t
\approx \vw^*$, $\calQ^t$ contains at least one true optimal configuration.
Finally, all options are required to be different in feature space. By
maximizing the utility of the objects, we are also pushing $\bbE[\langle \vw^t,
\Delta^t \rangle]$ towards zero, implying that iteration $t$ can only be
expected uninformative when $\vw^t$ is (approximately) anti-parallel to $\vw^*$:
$$ \bbE[u^*(\Delta^t)]  = 0 \iff \bbE[\langle \vw^*, \vw^t \rangle] \approx -\bbE[\norm{\vw^*}\norm{\vw^t}] $$
For reasonable users $\bbE[\langle \vw^*, \vw^t \rangle] = \bbE[\sum_{t=1}^T
u^*(\Delta^t)] \ge 0$ (by Lemma~\ref{th:ugaingeqzero}), implying that the above
case is extremely rare, and therefore $M \approx 0$.

This query strategy essentially attempts to find a good trade-off between
exploration ($\gamma \approx 1$) and exploitation ($\gamma \approx 0$). In most
cases a good strategy is to allow more exploration in the beginning of the
elicitation and then exploit more when the algorithm has learned a good
approximation of $\vw^*$. We therefore set $\gamma$ to $\frac{1}{t}$ in our
experiments. This also ensures that $u^t(\Delta^t)$ decreases over time
regardless of the user choice, thereby keeping $\beta$ constant.

In the following, we will stick to features $\vphi$ expressible as linear
functions of Boolean, categorical and continuous attributes. This choice is
very general, and allows to encode arithmetical, combinatorial and logical
constraints, as shown by our empirical evaluation.  So long as the feasible set
$\calY$ is also defined in terms of mixed linear constraint, query selection
can be cast as a mixed-integer linear problem (MILP) and solved with any
efficient off-the-shelf solver.

We remark that the previous arguments apply to all choices of $k \ge 2$, i.e.
to both pairwise and set-wise choice feedback. Intuitively, larger set sizes
imply more diverse and potentially more informative query sets, because they
reduce the chance for a reasonable user to pick a low utility option. They also
imply more conservative updates, mitigating the deleterious effect of
uninformative choices. These effects are studied experimentally.

\section{Empirical Evaluation}

We compare \cp\ against three state-of-the-art preference elicitation
approaches on three constructive preference elicitation tasks taken from the
literature. The query selection problem is solved with Gecode via its MiniZinc
interface~\cite{nethercote2007minizinc}\footnote{The complete experimental
setting can be retrieved from: \url{https://github.com/unitn-sml/choice-perceptron}}.

The three competitors are:
[i] the  Bayesian approach of~\cite{viappiani2010optimal} using Monte
Carlo methods (the number of particles was set to 50,000, as
in~\cite{teso2016constructive}) with greedy query selection based on the
Expected Utility of a Selection (a tight approximation of the Expected Value of
Information criterion);
[ii] Query Iteration, also from~\cite{viappiani2010optimal}, a sampling-based
query selection method that trades off query informativeness for computational
efficiency;
[iii] the set-wise maximum margin method of~\cite{teso2016constructive}, modified
to accept set-wise choice feedback; support for user indifference was
also disabled\footnote{These changes have no impact on the performance of the
method, and provide a generous boost to its runtime, due to the fewer pairwise
comparisons collected at each iteration.}.  We indicate the competitors as
\vbeus, \vbqi\ and \setmargin, respectively.  As argued in the previous section,
for \cp\ we set $\gamma$  to $\frac{1}{t}$ in all experiments, in order to allow
more exploration earlier on during the search.  In practice we also employ an
adaptive Perceptron step size, which is adapted at each iteration $t \ge 3$ from
the set $\{0.1, 0.2, 0.5, 1, 2, 5, 10\}$ via cross-validation on the collected
feedback; it was found to work well empirically.  \setmargin\ includes a similar
tuning procedure.

Our experimental setup is modelled after~\cite{teso2016constructive}. We
consider two different kinds of users: ``uniform'' and ``normal'' users, whose
true preference vectors $\vw^*$ are drawn, respectively, from a uniform and a
normal distribution. Twenty users are sampled at random and kept fixed for each
experiment. User responses are simulated with a Plackett-Luce
model~\cite{plackett1975analysis,luce1959individual}:
$$ P_x(\bar{y} = y_i | \calQ) = \frac{\exp(\lambda u^*(x, y_i))}{\sum_{j = 1}^k \exp(\lambda u^*(x, y_j))} $$
We set $\lambda=1$ as in~\cite{teso2016constructive}.  In the first two
experiments (which are context-less) we report the median over users of the
instantaneous regret, as in~\cite{viappiani2010optimal}
and~\cite{teso2016constructive}; whereas, in the third experiment (with context)
we report the median average regret.  In all experiments we also report
cumulative runtime and standard deviations.

\begin{figure*}[t]
    \begin{tabular}{cc|cc}
        \includegraphics[width=0.225\textwidth]{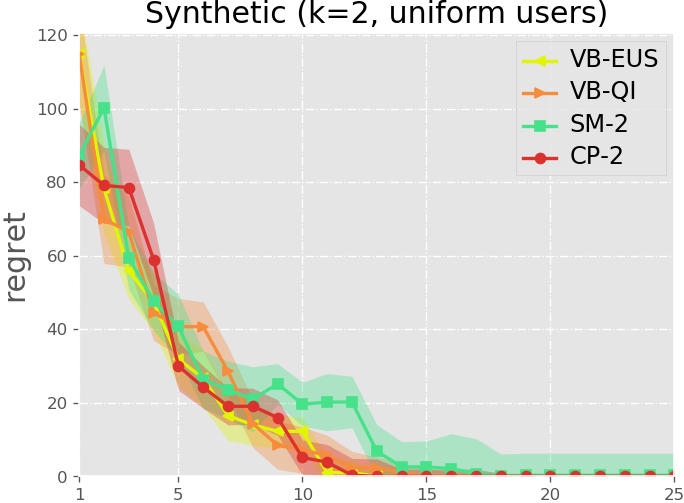} &
        \includegraphics[width=0.225\textwidth]{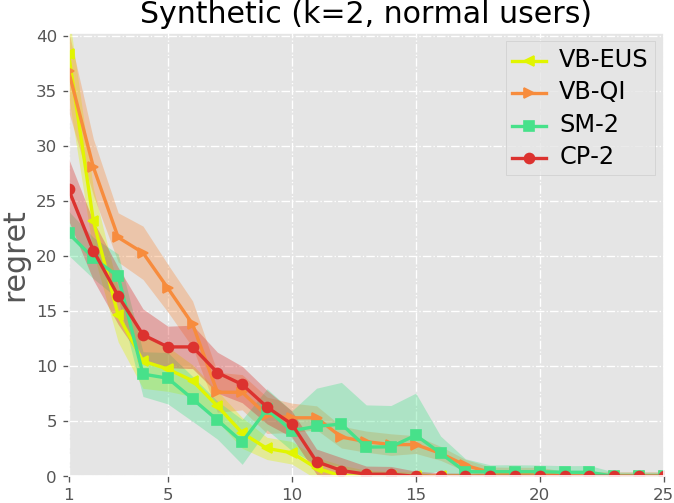} &
        \includegraphics[width=0.225\textwidth]{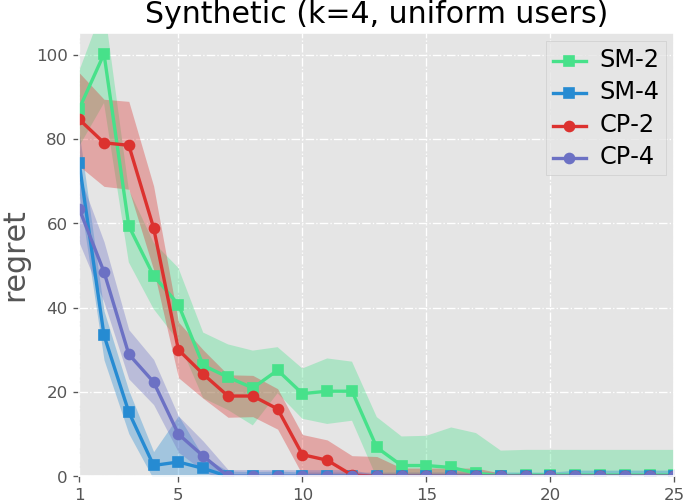} &
        \includegraphics[width=0.225\textwidth]{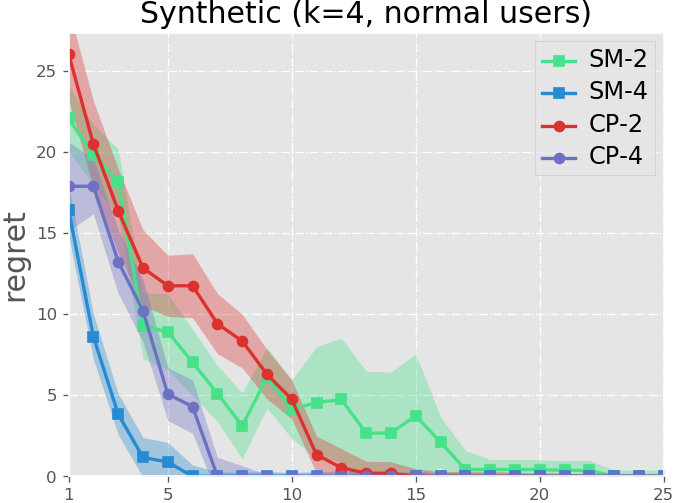}
        \\
        \includegraphics[width=0.225\textwidth]{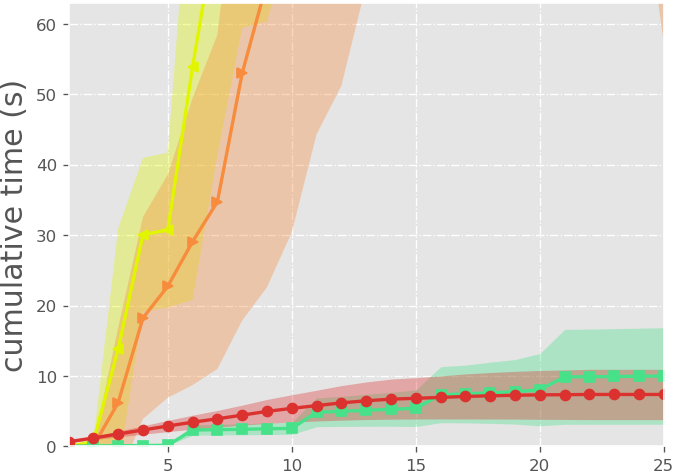} &
        \includegraphics[width=0.225\textwidth]{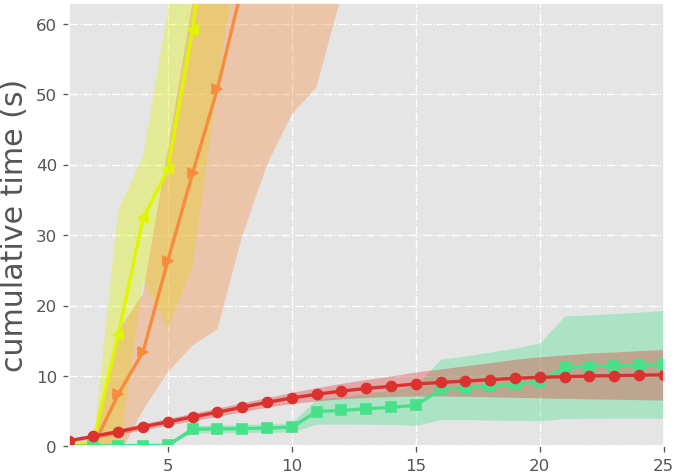} &
        \includegraphics[width=0.225\textwidth]{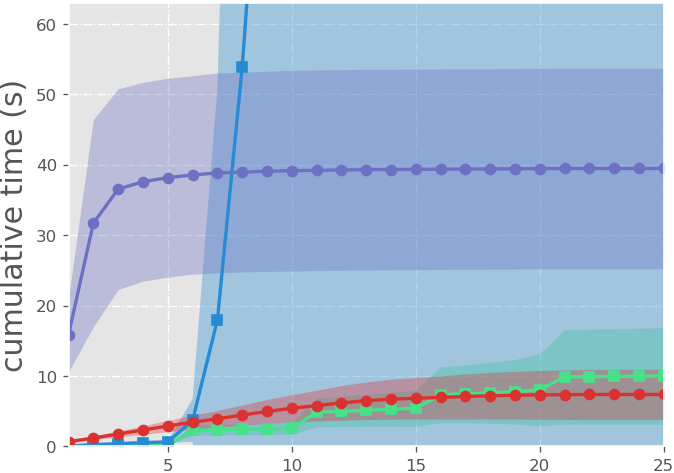} &
        \includegraphics[width=0.225\textwidth]{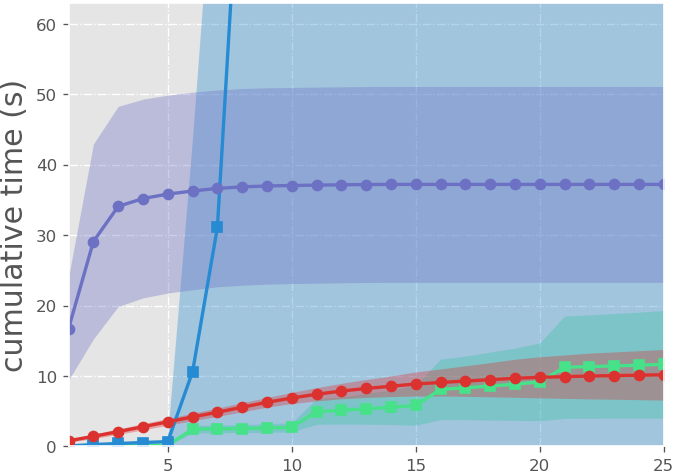}
    \end{tabular}
    \caption{\label{fig:synthetic} Comparison of various algorithms in the
synthetic experiment. The plots on the top row show the regret of the various
algorithm for increasing iterations, whereas the plots on the bottom row show
the cumulative running time (inference + learning). On the left, \cp\ is
compared to \setmargin, \vbeus and \vbqi using query sets with dimension $k=2$.
On the right, instead, \cp\ is compared only with \setmargin\ using $k=4$. In
both cases, experiments using uniformly distributed and normally distributed
users are shown on the left plots and on the right plots respectively. Best
viewed in color.
}
\end{figure*}

\begin{figure*}[t]
    \begin{tabular}{cc|cc}
        \includegraphics[width=0.225\textwidth]{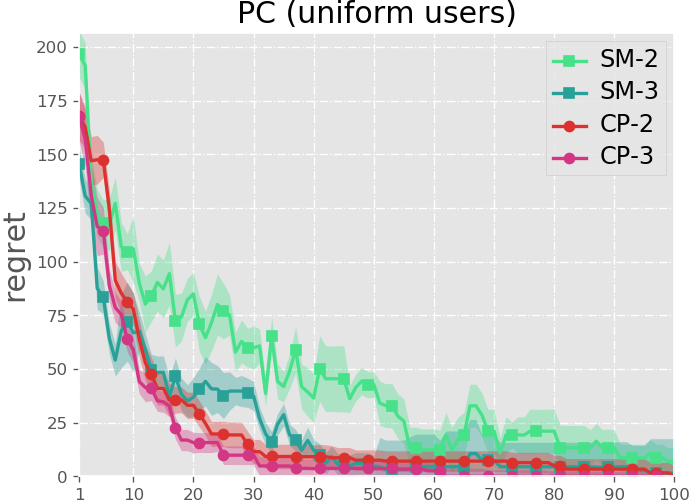} &
        \includegraphics[width=0.225\textwidth]{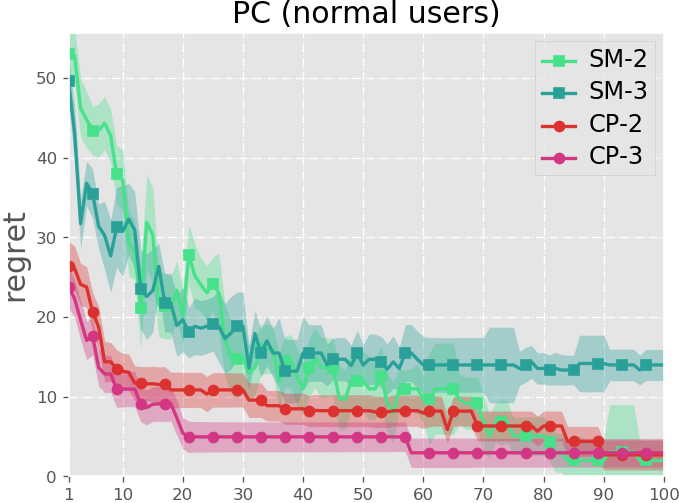} &
        \includegraphics[width=0.225\textwidth]{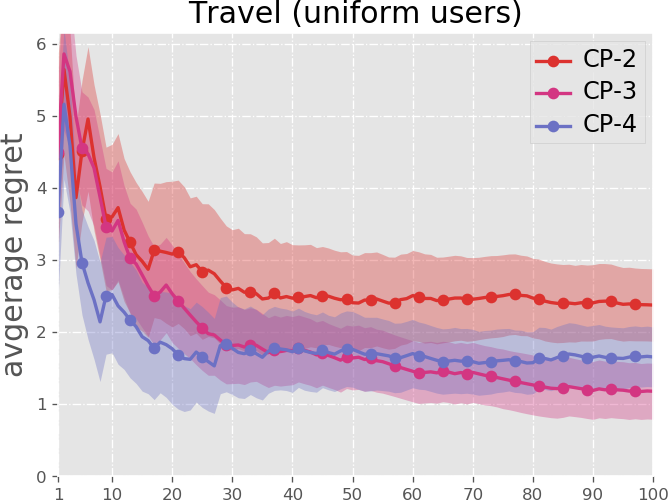} &
        \includegraphics[width=0.225\textwidth]{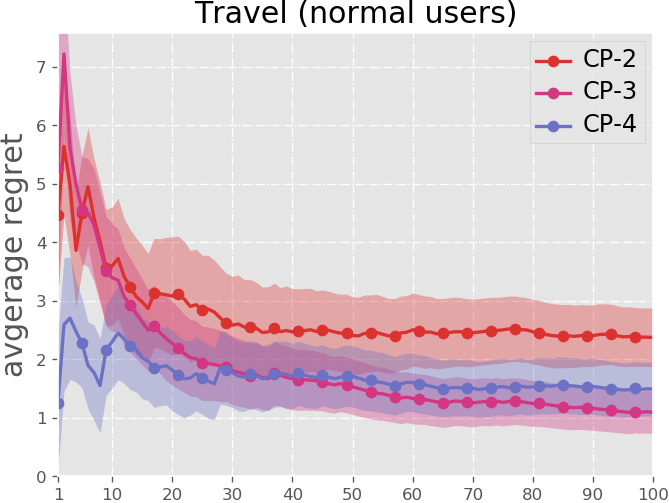}
        \\
        \includegraphics[width=0.225\textwidth]{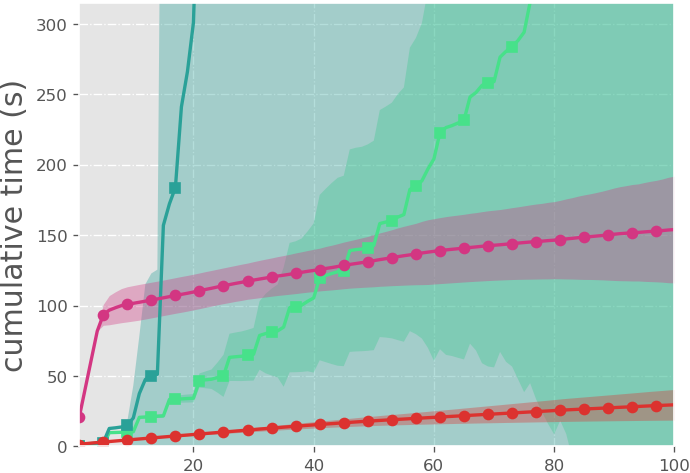} &
        \includegraphics[width=0.225\textwidth]{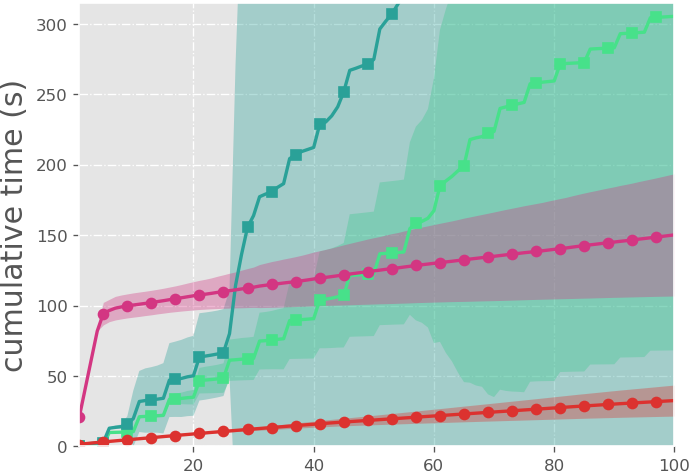} &
        \includegraphics[width=0.225\textwidth]{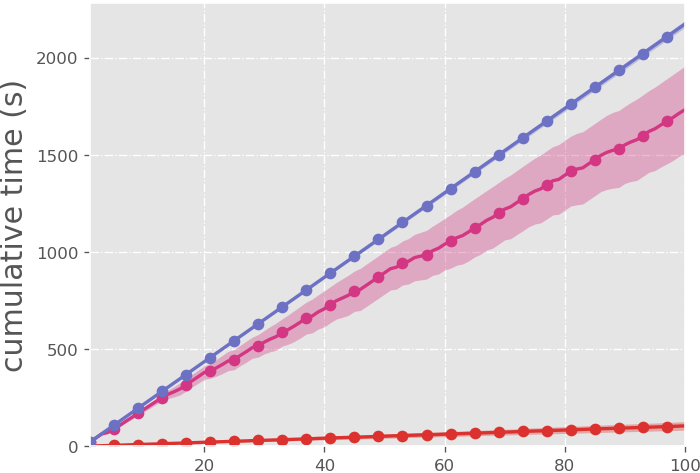} &
        \includegraphics[width=0.225\textwidth]{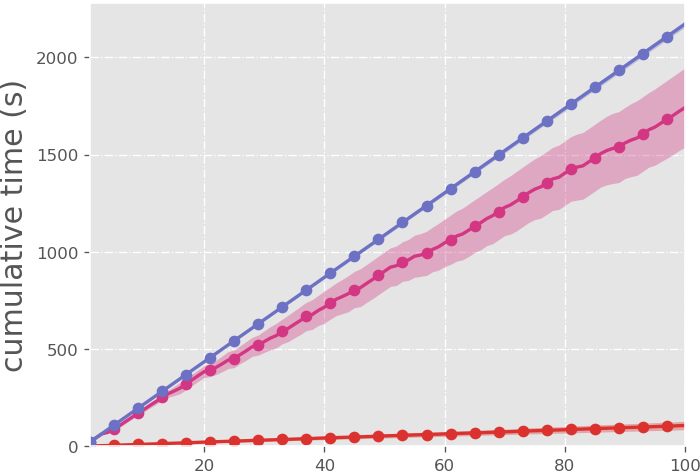}
    \end{tabular}
    \caption{\label{fig:pc} Comparison of various algorithms in the PC
configuration and travel planning experiments. The plots on the top row show the
regret of the various algorithm for increasing iterations, whereas the plots on
the bottom row show the cumulative running time (inference + learning). On the
left, \cp\ is compared to \setmargin\ on the PC configuration task using query
sets with dimension $k=2$ and $k=3$.  The plots on the right, instead, show only
the performance of \cp\ on the travel planning task using $k=\{2,3,4\}$. In both
cases, experiments using uniformly distributed and normally distributed users
are shown on the left plots and on the right plots respectively. Best viewed in
color.
}
\end{figure*}

\paragraph{Synthetic experiment.}
We evaluated all methods on the synthetic constructive benchmark introduced
in~\cite{teso2016constructive}. The space of feasible configurations is the
Cartesian product of $r$ attributes, each taking values in $[r]$, i.e.  $\calY =
\times_{i=1}^r [r]$. The features are the one-hot encoding of the attributes,
for a total of $r^2$ features. Here we focus on the $r = 4$ case ($16$ features,
$256$ products)
which is large enough to be non-trivial, and sufficiently small to be solvable
by the two Bayesian competitors. For \cp\ and
\setmargin\, $\calY$ is encoded natively via MILP constraints; the Bayesian
methods required $\calY$ to be enumerated. The users were sampled as
in~\cite{teso2016constructive}, i.e. from a uniform distribution in the range
$[1, 100]$ and a normal distribution with mean $25$ and standard deviation
$\frac{25}{3}$. All methods were run until either the user was satisfied (i.e.
the regret reported by the method reached zero) or 25 iterations elapsed.  We
evaluated the importance of the query set size by running \cp\ and
\setmargin\ with $k = 2, 3, 4$.  \vbeus\ and \vbqi\ were only run with $k = 2$,
due to scalability issues. In the $k = 2$ case (Figure~\ref{fig:synthetic},
left), \cp\ performs better than both \vbqi\ and \setmargin, and worse than
\vbqi.  The runtimes, however, vary wildly. The Bayesian competitors are much
more computationally expensive than \cp\ and \setmargin, confirming the
results of~\cite{teso2016constructive}; the two MILP methods instead avoid the
explicit enumeration of the candidate configurations, with noticeable
computational savings. Notably, \cp\ is faster than \setmargin, while
performing comparably or better. The gap widens with set size $k = 4$
(Figure~\ref{fig:synthetic}, right; $k = 3$ is similar, not shown). Here \cp\
and \setmargin\ converge after a similar number of iterations, but with
very different runtimes. The bottleneck of \setmargin\ is the
hyperparameter tuning procedure; disabling it however severely degrades the performance,
so we left it on.

\paragraph{PC configuration.}
In the second experiment, we compared \cp\ and \setmargin\ on a much larger
recommendation task, also from~\cite{teso2016constructive}. The goal is to
suggest a fully customized PC configuration to a customer. A computer is defined
by seven categorical attributes (manufacturer, CPU model, etc.) and a numerical
one (the price, determined by the choice of components).  The features include
the one-hot encodings of the attributes and the price. The relations between
parts (e.g. what CPUs are sold by which manufacturers) are expressed as Horn
constraints. The feasible space includes thousands of configurations, ruling the
Bayesian competitors out~\cite{teso2016constructive}. The users were sampled as
in the previous experiment. To help keeping running times low, the query
selection procedure of \cp\ is executed with a 20 seconds time cutoff. No time
cutoff is applied to \setmargin.

The results for $k=2$ and $3$ can be seen in Figure~\ref{fig:pc} (left). On
uniform users, \cp\ consistently outperforms \setmargin\ for both choices of
$k$, despite the timeout. Notably, \cp\ with $k=2$ (less informative queries)
works as well as \setmargin\ with $k=3$ (more informed queries) in this setting.
For normal users the situation is similar: with $k=2$, \setmargin\ catches up
with \cp\ after about 80 iterations, but at considerably larger computational
cost. Surprisingly, \setmargin\ behaves worse for $k=3$ than for $k=2$; \cp\
instead improves monotonically, for a modest increase in computational effort.
In all cases, the runtimes are very favorable to our method, also thanks to the
timeout, which however does not compromise performance.

\paragraph{Trip planning.}
Finally, we evaluated \cp\ on a slightly modified version of the touristic
trip planning task introduced in~\cite{teso2017coactive}. Here the recommender
must suggest a trip route between 10 cities, each annotated with
an offering of 15 activities (resorts, services, etc.).
The trip $y$ includes the path itself (which is allowed to contain
cycles) and the time spent at each city.
Differently from~\cite{teso2017coactive}, at each iteration the user issues a
context $x$ indicating a subset of cities that the trip must visit.  The
features include the number of days spent at each location, the number of times
an activity is available at the visited locations, the cost of the trip, etc.,
for a total of 127 features; see~\cite{teso2017coactive} for the details. Note
that this problem can not be encoded in \setmargin, i.e.  with Boolean and
dependent numerical attributes, without incurring significant encoding
overhead: the resulting \setmargin\ query selection problem would include
approximately 300 Boolean variables (an almost 300\% blow-up in problem size).
According to our tests, problems of this size are not solvable in real-time in
practice, compromising the reactiveness of \setmargin.

Differently from the previous two settings, here users were sampled from a
standard normal distribution (as in~\cite{teso2017coactive}) and from a uniform
distribution in the range $[-1, 1]$. Not having a one-hot encoded feature
vector, negative weights are useful to capture the user dislikes. The contexts
are uniformly sampled from the combinations of 2 or 3 cities. As in the previous
experiment, we employ a time cutoff of 20 seconds. We run this experiment with
$k={2,3,4}$ to show how different set sizes affect the performance of the
system. Since this experiment is context-based, we let the algorithm run for
exactly 100 iterations.  Figure~\ref{fig:pc} (right) reports the median average
regret and the median cumulative running time.

The plots show that in both cases there is a significant decrease in average
regret with $k=3$ over $k=2$, in exchange for increased running time; $k=4$
performs better than $k=3$ for about 40 iterations, but then worsens
considerably. This is probably due to the timeout, which in this more
complicated setting may substantially hinder the MILP solver. Increasing the
cutoff to 60 seconds however did not improve the results (data not shown).
This indicates that larger values of $k$ may be too costly to compute without
further approximations, as is also the case for the other competitors.

\paragraph{Choosing $k$}
While our theoretical analysis is agnostic on the
number $k$ of objects in a query set, in our empirical analysis we collected
some insight on how to choose $k$ on the basis of the difficulty of the
underlying optimization problem. While in general a larger $k$ is more
informative, it is not always possible to solve the query selection problem to
optimality. This may severely hinder the learning capabilities of the algorithm,
as in the case of the trip planning setting with $k=4$. On the other hand, for
smaller problems a larger $k$ may significantly reduce the number of iterations
needed to reach an optimal solution, as for the PC configuration setting. There
is, therefore, a trade-off that depends on the computational complexity of the
query selection problem of the application at hand. From our experiments, we can
infer, as a rule of thumb, that it is usually better to choose larger $k$
($k=4,5$) when objects are small and the selection problem easier to solve,
whereas a smaller $k$ ($k=2,3$) is preferable when the objects are large and
difficult to select. Additionally, the larger the objects, the harder it is for
the user to choose the best in the set, so a smaller $k$ is also desirable to
reduce the cognitive load on the user.

\section{Conclusion}

We presented the Choice Perceptron, an algorithm for preference elicitation
from noisy choice feedback. Contrary to existing recommenders, \cp\ can solve
constructive elicitation problems over arbitrary combinatorial spaces, composed
of many Boolean, integer and continuous variables and constraints. Our
theoretical analysis shows that, under a very general assumption, the average
regret suffered by \cp\ is upper bounded by $\calO(1/\sqrt{T})$.  The exact
constants appearing in the bound depend on intuitive properties of the query
selection strategy at hand. We further described a strategy that aims at
controlling these constants.
We applied \cp\ to constructive preference elicitation tasks for progressively
more complex combinatorial structures.  Not only \cp\ is the only method
expressive enough to deal with all of these problems, but it is also more
performant than the alternatives in terms of recommendation quality and
run-time.

In the future, we plan to research more informed query selection
strategies, e.g. by leveraging estimates of $\beta$ during query
selection. Other possible directions include exploring different
update rules. As mentioned, this algorithm and the analysis could be
also extended to perform exponentiated updates or handle generic
convex loss functions~\cite{shivaswamy2015coactive}.  Finally, a
deeper investigation on the optimal size of the query set and its
possible adaptation during the interaction process could be useful to
find an appropriate trade-off between informativeness and complexity.

\begin{small}
\bibliographystyle{aaai}
\bibliography{paper}
\end{small}

\end{document}